\documentclass{article}

\usepackage{arxiv}

\usepackage[utf8]{inputenc} 
\usepackage[T1]{fontenc}    
\usepackage{hyperref}       
\usepackage{url}            
\usepackage{booktabs}       
\usepackage{amsmath}
\usepackage{amsfonts}       
\usepackage{nicefrac}       
\usepackage{microtype}      
\usepackage{lipsum}		
\usepackage{graphicx}
\usepackage{pifont}
\usepackage[numbers,sort&compress]{natbib}
\usepackage{doi}

\newcommand{\cmark}{\ding{51}}%
\newcommand{\xmark}{\ding{55}}%

\title{Deep Learning on Edge TPUs}

\date{} 					

\author{ 
    {Yipeng Sun, }\href{https://orcid.org/0000-0003-3643-7776}{\includegraphics[scale=0.06]{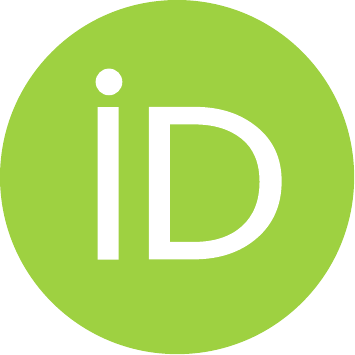}\hspace{1mm}Andreas M. Kist} \\
	Department Artificial Intelligence in Biomedical Engineering\\
	Friedrich-Alexander-University Erlangen-Nürnberg\\
	Germany \\
	\texttt{yipeng.sun@fau.de, andreas.kist@fau.de} 
}

\renewcommand{\headeright}{}
\renewcommand{\undertitle}{a survey}
\renewcommand{\shorttitle}{Deep Learning on Edge TPUs}

\hypersetup{
pdftitle={Deep Learning on Edge TPUs},
pdfsubject={q-bio.NC, q-bio.QM},
pdfauthor={Andreas M.~Kist},
pdfkeywords={Edge TPU, Deep Learning, Neural Network, Semantic segmentation},
}

\begin{document}
\maketitle

\begin{abstract}
		Edge computing is important in remote environments, however, conventional hardware is not optimized for utilizing deep neural networks. The Google Edge TPU is an emerging hardware accelerator that is cost, power and speed efficient, and is available for prototyping and production purposes. In this article, we review the Edge TPU platform, the tasks that have been accomplished using the Edge TPU, and which steps are necessary to deploy a model to the Edge TPU hardware. The Edge TPU is not only capable of tackling common computer vision tasks, but also surpasses other hardware accelerators, especially when the entire model parameters can be stored in the Edge TPU memory. Co-embedding the Edge TPU in cameras allows a seamless analysis of primary data. We further discuss future development directions, potential applications and limitations of the Edge TPU. In summary, the Edge TPU is a maturing system that has proven its usability across multiple computational tasks. 
\end{abstract}


\section{Introduction}

Deep neural networks (DNNs) have revolutionized image processing and computer vision, including object recognition, image classification and semantic segmentation \cite{lecun2015deep,sejnowski2018deep,garcia2017review}. These advances have led to significant paradigm changes, for instance in the field of healthcare \cite{esteva2019guide} and self-driving cars \cite{rao2018deep}. Different end-to-end deep learning platforms, such as Caffe, TensorFlow and Torch, are  important milestones in enabling the machine learning community to use deep neural networks. In particular, high-level packages, such as Keras and PyTorch, lowered the barriers to leveraging DNNs. Even with just a few lines of code, beginners can create, train, and evaluate for example a multilayer perceptron without the need for precise mathematical basics.

However, even with small DNNs the advantage of dedicated hardware, commonly graphical processing units (GPUs) is clear: faster training and inference, sometimes multiple orders of magnitude \cite{bergstra2011theano,wangBenchmarkingTPUGPU2019}. This becomes especially important in cases where DNNs are used in constrained environments, such as embedded solutions. Here, dedicated integrated circuits (ICs) will provide more efficient reasoning, i.e., more computation with less power and less latency, thus enabling real-time services. One idea is to utilize flexibly programmable field-programmable gate arrays (FPGAs) \cite{wang2016fpga,zhang2015fpga,shawahna2018fpga}, and another idea relies on dedicated ICs developed specifically for DNNs. Multiple platforms are already available, such as the NVIDIA Jetson family \cite{mittal2019survey}, Intel Movidius VPUs \cite{kaarmukilan2020accelerated} and Google's Edge TPUs \cite{cass2019taking} using a variety of frameworks \cite{hadidi2019characterizing}.

In this overview, we highlight the usage of Edge TPUs across deep neural network architectures and emphasize the authors' efforts in utilizing Edge TPUs on their specific tasks. We also discuss the future development, potential applications and limitations of Edge TPU.

\section{Overview of Deep Learning}
In order to explain the techniques discussed in this paper, we first provide a brief overview of deep learning. 

\begin{figure}[htbp]
	\centering
	\caption{DNN example of image classification. A picture of a cat is passed into a DNN used to identify dogs and cats, which finally yields a 0.96 probability of being a cat and a 0.04 probability of being a dog}
	\includegraphics[width=\linewidth]{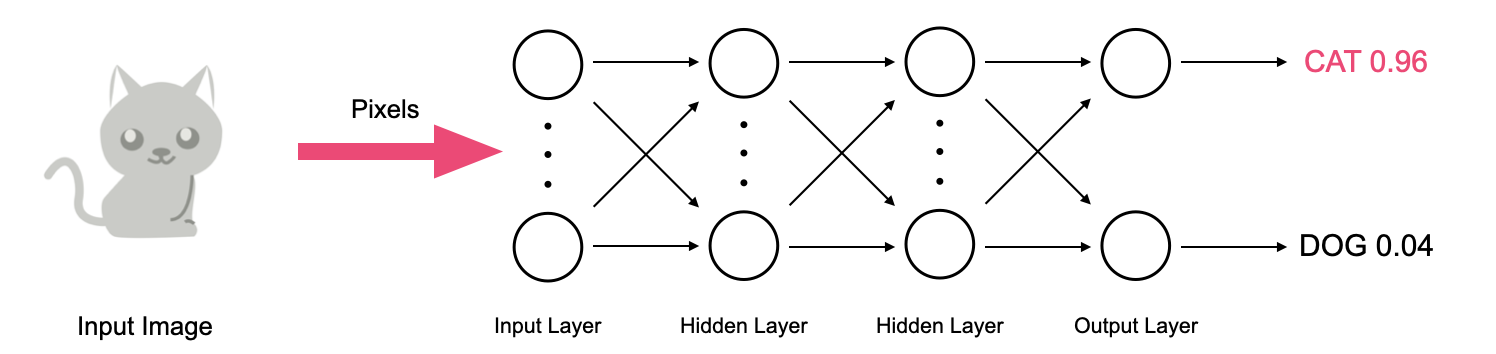}
	\label{fig:FCNN_structure}
\end{figure}

Deep learning prediction algorithms, are representational learning methods with multiple levels of representations, which is obtained by composing simple but nonlinear layers, as shown in Figure \ref{fig:FCNN_structure}. When the input data passes through the layers, each layer applies matrix operations to it. In most cases, the output of one layer is the input for the next. As a result of the final layer of data processing, a feature or classification output is generated. The neural network are called DNNs when they contains many layers in sequence. \cite{goodfellow2016deep}.

To optimize the parameters of matrix multiplications in each layer, the ground truth, also called training labels, are passed multiple times through the layers from the last layer to the first layer. That means in deep learning training, computations are reversed. A classical technique, Stochastic gradient descent (SGD) is usually used to minimize the error between prediction and ground truth, called training loss, in this case a small batch of samples is randomly selected and used to update the gradient according to the ground truth. With the continuous development of deep learning, more optimization algorithm are being developed. A single pass through the entire training dataset is referred to as a training epoch \cite{ruder2016overview}.

During the training of DNNs, numerous calculations are performed, resulting in DNNs models with a large number of parameters, which Leads to latency problems on the device. In addition, there are many choices of hyperparameters on how to design the DNNs model (e.g., number of parameters per layer and number of layers), causing the model design to be challenging. For example, DNNs with higher accuracy may require more memory, while DNNs with fewer parameters may run faster and use fewer computational resources, but can not meet the accuracy requirements of the practical application \cite{chen2019deep}. This problem was improved with the advent of hardware accelerators based on ICs, optimized specifically for deep learning algorithms, which provide a dramatic increase in the speed of running large-scale parametric models on these hardware accelerateors compared to conventional computing devices.

\section{The Edge TPU}

\begin{figure}[htbp]
	\centering
	\caption{Supported operations by the Edge TPU (black) and operations with known limitations (gray). Dates were retrieved from Wayback Machine and are not actual updates by Coral.}
	\includegraphics{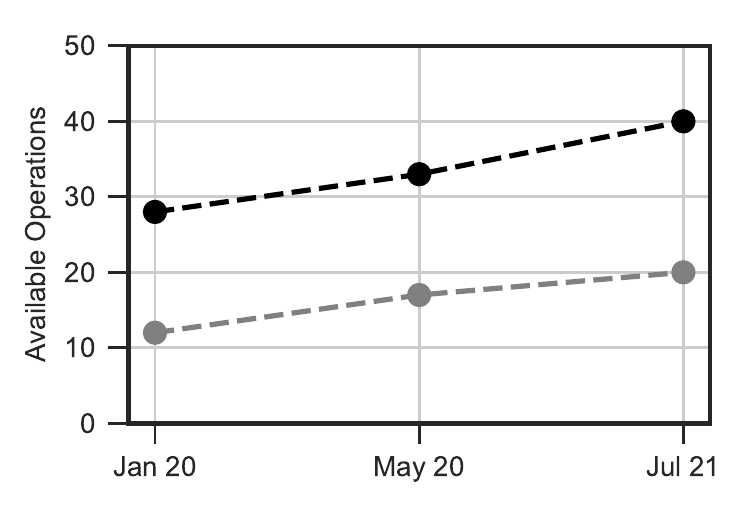}
	\label{fig:supportedOperations}
\end{figure}

The Edge TPU is a new machine learning application-specific integrated circuit (ASIC) with a small footprint of 5 $\times$ 5 mm from Google. It allows fast TensorFlow Lite model inference at low power. The available operations on the Edge TPU are constantly growing \cite{coralOperations}, and are updated regularly. The number of operations currently available (July 2021) is shown in figure \ref{fig:supportedOperations}. Interestingly, over time the ratio of the number of operations with known limitations to all operations supported by the Edge TPU remains constant by approximately 50 \%. Among the recent updates, it is worth mentioning that the introduction of long and short-term memory (LSTM) units has enabled the use of recurrent neural networks (RNNs) on the Edge TPU.

With a power consumption of only two watts, the Edge TPU is well suited for low power environments. Specifically, on selected deep neural networks the Edge TPU outperforms other hardware accelerators when measuring images per second per Watt \cite{kljucaric_architectural_2020}. Each Edge TPU IC can operate at four trillion operations per second (TOPS), resulting in two TOPS per Watt. Currently, multiple hardware platforms are available for prototyping and production purposes (Table \ref{tab:edgetpus}). 

\begin{table}[htbp]
	\caption{Systems for Edge TPU inference}
	\centering
    \begin{tabular}{lrlrr}
    \hline
    product   & purpose & vendor &  TOPS & est. costs \\
    \hline
    Dev Board & Prototyping   & Coral / Google  &  4 & 129.99 USD \\
    Dev Board Mini & Prototyping   & Coral / Google  & 4 &  99.99 USD \\
    Dev Board Micro & Prototyping   & Coral / Google  & 4 &  79.99 USD \\
    USB Accelerator & Prototyping   & Coral / Google  &  4 & 59.99 USD \\
    Mini PCIe Accelerator & Production   & Coral / Google  &  4 & 24.99 USD \\
    M.2 Accelerator (A+E or B+M key) & Production   & Coral / Google  & 4 &  24.99 USD \\
    M.2 Accelerator (Dual Edge TPU) & Production   & Coral / Google  & 8 &  39.99 USD \\
    System-on-Module (SoM) & Production   & Coral / Google  & 4 &  99.99 USD \\
    AI Accelerator PCIe Card (8/16 Edge TPUs) & Production   & Asus  & 32/64 &  1200 USD (8) \\
    Accelerator Module (IC) & PCBs   & Coral / Google  & 4 &  19.99 USD \\
    
    \hline
    \end{tabular}
	\label{tab:edgetpus}
\end{table}

\section{Computing on the Edge}
\label{sec:headings}

\begin{figure}[htbp]
	\centering
	\caption{Tasks accomplished using the Edge TPU}
	\includegraphics[]{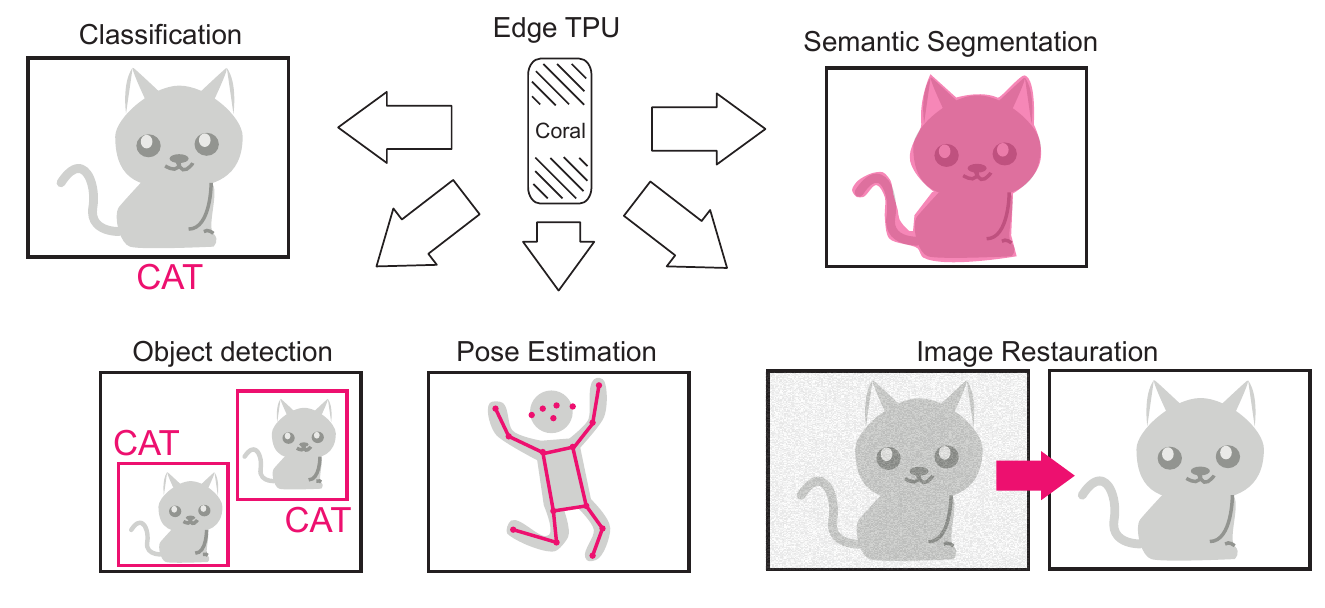}
	\label{fig:edgetpu}
\end{figure}

With the Coral platform, Google provides a comprehensive library of models that can be directly utilized. Figure \ref{fig:edgetpu} shows the tasks that were successfully completed using the Edge TPU. In the following, we will take a closer look to each of the tasks. In general, recent work has focused on comparing the Edge TPU with other hardware accelerateors in terms of accuracy, inference time, and power consumption. Specifically, they focus on the compatibility of Edge TPUs with existing architectures. A comprehensive approach was provided in \cite{kong_edlab_2021}. The authors suggest a new benchmark (EDLAB) to compare hardware accelerators and show that the Edge TPU is leading in power consumption and is essentially equal in accuracy compared to other hardware accelerators. Interestingly, in their study the Edge TPU is slower in inference compared to its competitors, which is contrary to most observations below. Another study performed a bottleneck analysis of the Edge TPU, tested 24 neural networks, and found three shortcomings of the Edge TPU in terms of computational throughput, energy efficiency, and memory access handling. Aiming for improving these issues, they proposed a new framework, Mensa, that improves energy efficiency and throughput by 3.0x and 3.1x over the Edge TPU. It is promising that Mensa could enable the design and adoption of future heterogeneous accelarators, especially Edge TPUs, to support the type of DNNs models that have not yet been developed.\cite{boroumand_mitigating_2021}. 

\subsection{Classification}

In \cite{wisultschew_artificial_2019}, the authors tested Edge TPUs with multiview convolutional neural networks \cite{su_multi-view_2015} to classify objects from various views and specialized convolutional neural networks. They found that the Edge TPU outperformed the CPU, ARM CPU and the Intel Movidius NCS by a large margin. The Edge TPU had a latency of less than 5 ms, and was highly power efficient providing 451.8 forward passes per Watt per second \cite{wisultschew_artificial_2019}. However, the authors did not compare the accuracy of the Edge TPU model (uint8) to the classification accuracy of the competing platforms (float32). In \cite{asyraaf_jainuddin_performance_2020}, the authors compared different neural architectures for object classification deployed to the Edge TPU. Their results suggest that the combination of Raspberry Pi 3 or 4 with the Edge TPU is feasible across different architectures, allowing an increase in the number of processed frames per second without significant loss in accuracy. The use of Siamese networks on Edge TPUs has been shown in \cite{lungu_siamese_2020}, where the authors showed that Edge TPUs are capable of providing inferences at 60 frames per second, and quantization does not hinder the performance. Notably, the authors observed a small increase in performance. The analysis of audio spectrograms on Edge TPUs was assessed by \cite{hosseininoorbin_scaling_2021}, where the authors showed that the Edge TPU is on par in accuracy with CPU-based deep neural networks. However, they also found that the power efficiency of Edge TPUs is superior to CPUs only when the complete set of model parameters are deployed to the Edge TPU \cite{hosseininoorbin_scaling_2021}. A direct comparison of NVIDIA Jetson Nano and the Edge TPU revealed that architectures that cache their complete parameter set to the Edge TPU SRAM are approximately five times faster (peak inference 417 frames per second) than their Jetson Nano counterpart \cite{kang_benchmarking_2021}. In agreement, architectures with uncached parameters were of the same order as the Jetson Nano. In summary, Edge TPU has a great speed advantage over traditional embedded computing processing units in processing classification tasks in remote computing scenarios. However, in the above experiments, they did not compare the accuracy of the Edge TPU model with int8 precision and competitors with float32 precision.

\subsection{Object detection}

In the COVID-19 pandemic, the automatic detection of face masks is important for the prevention of infections. The authors of \cite{park_real-time_2020} showed that a single-shot detector (SSD) \cite{liu2016ssd} with a MobileNetV2 backbone \cite{sandler2018mobilenetv2} is portable to the Edge TPU to classify detected faces for mask and no-mask in 6.4 ms per frame. A similar configuration was reported in \cite{puchtler_evaluation_2020}. It is worth mentioning that, the authors tested the MobileNetV2-SSD architecture on the MS COCO dataset and found that the mean average precision (mAP) was for the Edge TPU quantized variant 0.2248, only slightly lower than the float32 competitors (Raspberry Pi 3 mAP 0.2530, with Intel Movidius NCS 0.2459). However, the Edge TPU surpassed the competitors in inference time by providing 55 frames per second. An SSD has been additionally used with MobileNetV1 and MobileNetV2 to analyze wine trunks, where the Edge TPU served as the acceleration module \cite{aguiar_visual_2020}. The average inference time across MobileNets was approximately 20-24 ms, which the authors compared to Tiny YOLO-V3 on an NVIDIA Jetson platform (54 ms). The Edge TPU SSDs also outperformed the Tiny YOLO-V3 in terms of average precision. 

\subsection{Pose estimation}

Real-time pose estimation is a feasible approach in the training of pose-relevant sports to provide immediate feedback to the trainee. In a recent study, the Edge TPU was used to extract pose information from golfing footage \cite{kimApplyingPoseEstimation2020}. Using a combination of the Edge TPU prediction and a Savitzky-Golay filter, the authors were able to obtain a pose estimation accuracy of up to 81.2 \%. Edge TPUs have recently been used for 3D pose estimation \cite{bultmannRealTimeMultiView3D2021}. The authors utilized the Edge TPU to allow real-time 3D pose estimation of three persons in a single camera frame at 30 Hz. In detail, each image crop took approximately 4.5 ms, and in each second the Edge TPU-based object detection took another 20 ms. 

\subsection{Image denoising}

The authors of \cite{abeykoon_scientific_2019} ported TomoGAN \cite{liu_tomogan_2020}, a generative adversarial network, to the Edge TPU to denoise X-ray images at the edge. The authors highlighted that quantizing the model decreases the structural similarity index (SSIM) \cite{wang_image_2004}. However, the SSIM can be rescued by fine-tuning the model \cite{abeykoon_scientific_2019}.

\subsection{Semantic segmentation}

The first description of using Edge TPUs for semantic segmentation was in \cite{kist2020efficient}. The authors mined different segmentation architectures: the SegNet \cite{badrinarayanan2017segnet}, the DeepLabV3+ \cite{chen2017rethinking} and the U-Net \cite{ronneberger2015u} architecture. Specifically, by dynamically scaling the U-Net architecture, smaller U-Net derivatives can be completely deployed by maintaining high segmentation accuracy \cite{kist2020efficient}. When using large images, the resizing operations on the Edge TPU are mapped to the CPU to avoid precision loss \cite{coralOperations}. However, this yields the fact that not all operations are fully mapped to Edge TPU resulting in slow inference speeds. By using a custom upsampling algorithm consisting of tiling, upsampling and merging, the authors showed that they were able to fully deploy the network to the Edge TPU, dramatically increasing the throughput without a significant drop in segmentation accuracy \cite{kist2020efficient}.

\section{From TensorFlow/Keras to Edge TPU}

\subsection{Porting existing and implementing novel architectures}

As described in the previous sections, many works have ported existing architectures to the Edge TPUs. However, due to the lack of adequate operations, the one-to-one porting is sometimes neither possible \cite{kist2020efficient} nor desired, as shown in \cite{gupta2020accelerator}. The authors showed that an Edge TPU-specific version of the EfficientNet-family that uses ordinary convolutions and the ReLU activation function is faster and more accurate compared to the separable convolutions and the swish activation function typically used in EfficientNets \cite{tan_efficientnet_2019}. \cite{yazdanbakhsh_evaluation_2021} developed a model that estimates the latency of an architecture deployed to the Edge TPU with a high accuracy of 97 \%. For this model, the authors utilized more than 423,000 unique convolutional neural networks (CNNs) leading to a reliable estimate to allow rapid evaluation of architectural design choices. Accelerator-aware optimization is therefore key for future DNN architectures that are efficient on the Edge TPU.

Table \ref{tab:appl} illustrates that most networks available through \verb+tf.keras.applications+ (TensorFlow 1.15 as recommended by Coral) can be easily deployed to the Edge TPU without any further modifications. The code used to generate this table is available at \url{https://github.com/anki-xyz/edgetpu}. The Edge TPU compiler in version 16 was used in the Google Colab environment. In detail, MobileNets are the only available networks through \verb+tf.keras.applications+ that are completely mapped to the Edge TPU and do not occupy the full Edge TPU SRAM allowing fast inference. The NASNet family, as well as the Xception architecture have unsupported subpaths early in the graph resulting in limited mapping of operations to the Edge TPU. 

\begin{table}[htbp]
	\caption{Deep neural networks ready to deploy to Edge TPU. OPs = operations}
	\centering
    \begin{tabular}{lrlrr}
    \hline
    model   & SRAM full &  Total OPs &  OPs on Edge TPU \\
    \hline
    MobileNet \cmark         &        no &                39 &                             39 \\
    MobileNetV2 \cmark       &        no &                72 &                             72 \\
    DenseNet121       &       yes &               384 &                            384 \\
    DenseNet169       &       yes &               533 &                            533 \\
    DenseNet201       &       yes &               622 &                            622 \\
    InceptionResNetV2 &       yes &               395 &                            395 \\
    InceptionV3       &       yes &               162 &                            162 \\
    NASNetLarge \xmark       &        no &              1331 &                              2 \\
    NASNetMobile \xmark      &        no &               991 &                              1 \\
    ResNet101         &       yes &               145 &                            145 \\
    ResNet101V2       &       yes &               249 &                            249 \\
    ResNet152         &       yes &               213 &                            213 \\
    ResNet152V2       &       yes &               368 &                            368 \\
    ResNet50          &       yes &                77 &                             77 \\
    ResNet50V2        &       yes &               130 &                            130 \\
    VGG16             &       yes &                24 &                             24 \\
    VGG19             &       yes &                27 &                             27 \\
    Xception \xmark          &        no &               128 &                             11 \\
    \hline
    \end{tabular}
	\label{tab:appl}
\end{table}

\subsection{Deployment}

\begin{figure}[htbp]
	\centering
	\caption{Deployment of neural networks to the Edge TPU. Asterix denotes that models should either be trained in quantization aware mode or quantized after conventional training.\vspace{20pt}}
	\includegraphics[width=\linewidth]{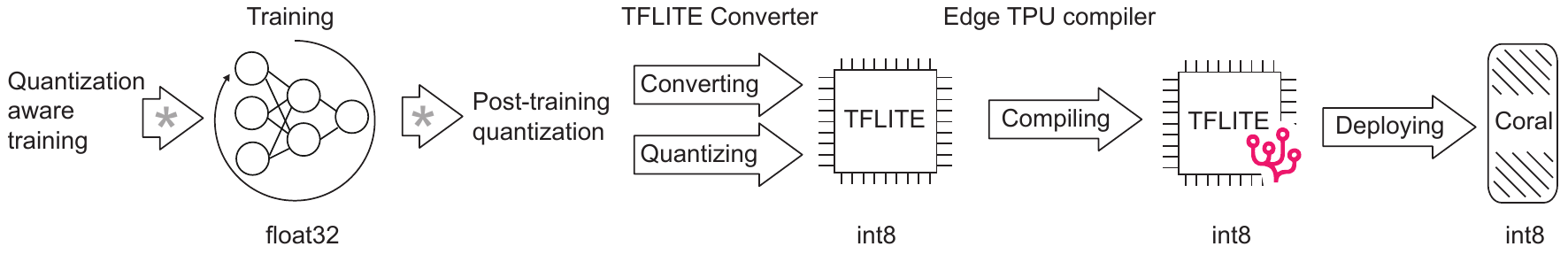}
	\label{fig:deployment}
\end{figure}

The deployment of DNNs to the Edge TPU is a multistep process (Figure \ref{fig:deployment}). First, a TensorFlow or Keras model is converted to the \verb+TFLITE+ format. As the model will be quantized (int8 or uint8), the model should be either trained in quantize-aware mode or post-training quantized, where the former is preferred for the best performance. In this step, it may be necessary to provide a representative dataset for effective quantization. Next, the Edge TPU compiler uses the \verb+TFLITE+ file to compile it into a special Edge TPU \verb+TFLITE+ format, deciding which operations are mapped to the Edge TPUs or to the CPU. Even though all operations can be mapped to the Edge TPU, it may happen that weights and parameters have to be transferred to the Edge TPU during inference because of the limited SRAM of the Edge TPU (see also Table \ref{tab:appl}). Although deployment and inference are platform independent, the Edge TPU compiler needs a UNIX environment. However, one can utilize the Google Colab platform to compile \verb+TFLITE+ files.
Typically, the purpose of the Edge TPU is inference. Nevertheless, there are two options to retrain the DNN on the Edge TPU to allow transfer learning \cite{transferLearning}. One method is the retraining of the last layer using backpropagation and cross entropy. The other method uses weight imprinting. Here, the embedding of the pre-last layer is used to compute the new weights in the final model layer.

\section{Embedded Edge TPUs}

Edge TPUs can be easily incorporated on PCBs and interfaced with PCIe and USB 2.0. Multiple companies already included Edge TPUs in network-attached storage (NAS)\footnote{\url{https://www.qnap.com/en/product/g650-04686-01}} or telematic solutions such as network switches\footnote{\url{https://www.nexcom.com/Products/mobile-computing-solutions/ai-edge-telematics-solution/google-coral-solution}}. A very promising application is the integration of Edge TPUs in cameras. These so-called smart cameras allow the direct processing of acquired images and can provide not only the raw image but also direct information about what is in the image. Table \ref{tab:cameras} provides an overview of commercially available cameras that contain an Edge TPU. The only system that has an industry standard objective mount (C-Mount) is the Vision AI platform. The MP Cam and JeVois cameras are prototyping platforms, whereas the Darcy camera comes in a production-friendly package.

\begin{table}[htbp]
	\caption{Smart cameras with Edge TPUs}
	\centering
	\begin{tabular}{lllll}
		\hline
		product     & vendor     & camera chip & resolution [px] & est. cost \\
		\hline
		MP Cam      & Siana Systems      & OV5640    & 2592$\times$1944 (5 MP) & 275 USD    \\
		Vision AI   & IMAGO Technologies & -         & 2560$\times$1936 (5 MP) & 1500 USD  \\
		Darcy       & edgeworx           & -         & 1024$\times$768 (0.8 MP) & not known \\
		JeVois      & JeVois Inc.        & -         & 1280$\times$1024 (1.3 MP) & 50 USD \\
		\hline
	\end{tabular}
	\label{tab:cameras}
\end{table}

\section{Discussion}

\subsection{Edge TPUs in Computer Vision}

As Figure \ref{fig:edgetpu} shows, the current focus and therefore, main application of Edge TPUs is the analysis of images. When deploying DNNs to the Edge TPU, one may observe performance drops that can be rescued by adjusting the architecture \cite{kist2020efficient} or fine-tuning the model \cite{abeykoon_scientific_2019}. As the Edge TPU is one hardware accelerators out of many, it is important to compare it to other candidates \cite{aguiar_visual_2020,kang_benchmarking_2021,wisultschew_artificial_2019}, to identify if the Edge TPU is the most appropriate technological platform. Standardized benchmarks, as suggested by \cite{kong_edlab_2021}, for different computer vision tasks, or for identifying power efficiency are crucial to ensure fair comparisons across hardware accelerators. 
Across most of the studies, the Edge TPU outperformed non-accelerating platforms (CPU, Raspberry Pis) and other hardware accelerators. However, it has to be noted that other hardware platforms have an easier deployment due to the richer repertoire of supported operations (NVIDIA Jetson family supports native TensorFlow) and operate not only in int8 precision. Here, both main competitors, NVIDIA Jetson and Intel Neural Compute Stick 2, are able to use not only int8, but also floating point 16 (FP16) or even FP32 precision (NVIDIA Jetson) as well. This additional accuracy could be crucial on selective tasks, such as identity verification. 

\subsection{Future applications}

The Edge TPU is an embedding-friendly hardware accelerator. Therefore, in future we expect to see more Internet of Things (IoT) devices that will incorporate the Edge TPU. A potential application that has not been touched is the deployment of DNNs with multimodal input, such as videos that contain both, audio and image signals \cite{tzirakis2017end}. This is potentially interesting for intelligent vehicles relying on multiple sensors, or medical assessment. 
We envision that an ideal application for the Edge TPU is the pure inference of large datasets, for example for data categorization by providing high-level embedding \cite{frome2013devise}. Furthermore, embedding the Edge TPU together with adjacent hardware, such as cameras, is highly promising and the first approaches are already commercially available. This close proximity to the camera hardware allows a manifold of image prepossessing, such as image denoising and restoration, super-resolution applications, and camera setting adjustments. Ideally, the Edge TPU allows the direct analysis of primary data and only transmits the inference results to downstream receivers.

\section{Conclusion}

The Edge TPU is a powerful platform, especially for accelerating inference on the edge and in remote settings. Despite its major usage in the image analysis domain (Figure \ref{fig:edgetpu}), DNNs that integrate multiple sources of data are possible. Although the Edge TPU has limitations in supported operations \cite{coralOperations} and is limited to int8 precision, with architecture adjustments \cite{kist2020efficient} and fine-tuning \cite{abeykoon_scientific_2019}, one is able to achieve competitive results. 

\section*{Acknowledgment}

Thanks Michael Döllinger, Tobias Schraut and René Groh for their critical comments on the manuscript.

\section*{Declarations}

\subsection*{Funding}

N/A

\subsection*{Conflicts of interest/Competing interests}

The authors declare no competing interests.

\subsection*{Availability of data and material}

All primary data is available at \url{https://github.com/anki-xyz/edgetpu}.

\subsection*{Code availability}

All code to generate primary data is available at \url{https://github.com/anki-xyz/edgetpu}.

\subsection*{Authors' contributions}

Andreas M. Kist and Yipeng Sun conceived, wrote and edited the article, generated primary data and created figures and tables.

\bibliographystyle{IEEEtranN}
\bibliography{deeplearningedge}  

\end{document}